\documentclass[sigconf]{acmart}

\usepackage{amsthm}
\usepackage{algorithm}
\usepackage{algorithmic}
\usepackage{multirow}

\AtBeginDocument{%
  }

\setcopyright{acmlicensed}
\copyrightyear{2018}
\acmYear{2018}
\acmDOI{XXXXXXX.XXXXXXX}

\acmConference[Conference acronym 'XX]{Make sure to enter the correct
  conference title from your rights confirmation email}{June 03--05,
  2018}{Woodstock, NY}
\acmISBN{978-1-4503-XXXX-X/18/06}

\begin{document}

\title{LMP: Leveraging Motion Prior in Zero-Shot Video Generation \\ with Diffusion Transformer}

\author{Changgu Chen}
\affiliation{%
  \institution{East China Normal University}
  \country{}
}
\orcid{0009-0007-2219-7069}

\author{Xiaoyan Yang}
\affiliation{%
  \institution{East China Normal University}
  \country{}
}
\orcid{0009-0004-9179-2185}

\author{Junwei Shu}
\affiliation{%
  \institution{East China Normal University}
  \country{}
}

\author{Changbo Wang}
\affiliation{%
  \institution{East China Normal University}
  \country{}
}
\orcid{0000-0001-8940-6418}
\authornote{Corresponding authors. Email: \{cbwang, yli\}@cs.ecnu.edu.cn}

\author{Yang Li}
\affiliation{%
  \institution{East China Normal University}
  \country{}
}
\orcid{0000-0001-9427-7665}
\authornotemark[1]

\renewcommand{\shortauthors}{Chen et al.}

\begin{abstract}
In recent years, large-scale pre-trained diffusion transformer models have made significant progress in video generation. While current DiT models can produce high-definition, high-frame-rate, and highly diverse videos, there is a lack of fine-grained control over the video content. Controlling the motion of subjects in videos using only prompts is challenging, especially when it comes to describing complex movements. Further, existing methods fail to control the motion in image-to-video generation, as the subject in the reference image often differs from the subject in the reference video in terms of initial position, size, and shape. To address this, we propose the \textbf{L}everaging \textbf{M}otion \textbf{P}rior (\textbf{LMP}) framework for zero-shot video generation. Our framework harnesses the powerful generative capabilities of pre-trained diffusion transformers to enable motion in the generated videos to reference user-provided motion videos in both text-to-video and image-to-video generation.
To this end, we first introduce a foreground-background disentangle module to distinguish between moving subjects and backgrounds in the reference video, preventing interference in the target video generation. A reweighted motion transfer module is designed to allow the target video to reference the motion from the reference video. To avoid interference from the subject in the reference video, we propose an appearance separation module to suppress the appearance of the reference subject in the target video.
We annotate the DAVIS dataset with detailed prompts for our experiments and design evaluation metrics to validate the effectiveness of our method. Extensive experiments demonstrate that our approach achieves state-of-the-art performance in generation quality, prompt-video consistency, and control capability.
\end{abstract}




\begin{teaserfigure}
  \includegraphics[width=\textwidth]{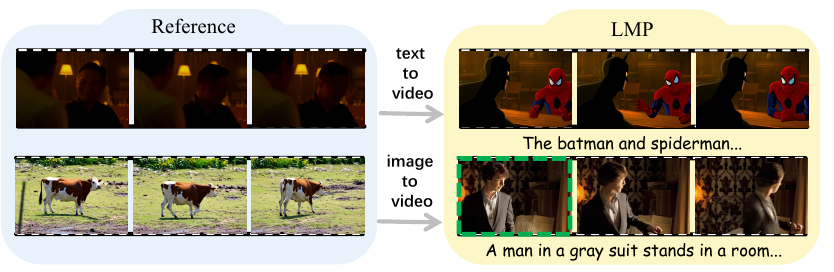}
  \caption{Our LMP framework enables DiT-based video generation models to produce target videos that reference the motion from a reference video in both text-to-video and image-to-video settings. The green dashed box indicates the reference image.}
  \label{fig:teaser}
\end{teaserfigure}


\received{20 February 2007}
\received[revised]{12 March 2009}
\received[accepted]{5 June 2009}

\maketitle

\section{Introduction}
Diffusion-based generative models \cite{ho2020denoising,song2020denoising,rombach2022high} have demonstrated impressive performance across various tasks, including text-to-image \cite{saharia2022photorealistic,ramesh2022hierarchical,rombach2022high}, text-to-video \cite{guo2023animatediff,wang2023modelscope}, and text-to-3D generation \cite{liu2023zero,voleti2024sv3d}. In recent years, the architecture of diffusion models has evolved from the original UNet \cite{ho2020denoising,song2020denoising,rombach2022high} to the current DiT \cite{flux2024, yang2024cogvideox} architecture. DiT-based video generation models have made significant progress compared to UNet-based architectures. They have achieved better video clarity, longer video duration, and greater semantic diversity. However, relying only on prompts makes it difficult to control the fine-grained motion details in the generated videos. Leveraging motion priors to enhance fine-grained motion control in generated videos remains a significant challenge.

There are currently two primary categories to address this challenge. The first is training-based approaches. While ControlNet-like methods~\cite{zhang2023adding,zhao2024uni,qin2023unicontrol,mou2023t2i} have introduced depth maps or edge maps to precisely control motion in generated videos, They are limited by control signals that are difficult to obtain. For instance, these methods struggle to handle varying sizes, positions, and shapes of moving objects in the videos. Moreover, obtaining the desired additional control signals is also a challenge for users. Other methods~\cite{zhao2024motiondirector,wang2024motionctrl} train networks to adapt to additional motion signals, such as trajectories, bounding boxes, and motion masks. However, these approaches require extensive training resources and may suffer from biases due to limited fine-tuning data, leading to suboptimal performance on rare motion patterns.
The second is zero-shot approaches~\cite{chen2024motion,jain2024peekaboo}. These methods cleverly leverage the attention properties of UNet-based diffusion models to control the trajectories and poses of moving objects. However, such methods heavily rely on the architecture of the base model and are not universally applicable to DiT architectures. Further, previous motion control methods were designed only for text-to-video and did not adapt to the image-to-video setting. The position, size, and shape of the subject in the reference image often differ from those in the reference video. This makes it difficult to directly apply previous methods to the image-to-video setting. Therefore, there is a need to design a method that can leverage the properties of the latest DiT model architecture to control the motion of subjects without requiring training, and is applicable to both text-to-video and image-to-video settings.

In Unet-based diffusion video generation models, there are three types of attention layers: \textbf{self-attention}, which primarily controls content within frames; \textbf{cross-attention}, which ensures consistency between generated content and prompts; and \textbf{temporal-attention}, which manages dynamics between frames. In contrast, the latest DiT architecture unifies different modalities, conditions, and spatiotemporal information into tokens, processing them all with a single self-attention mechanism \cite{yang2024cogvideox}. This allows each token in the DiT architecture to have a global view, accessing all information, and making it possible to reference motion from another video. Furthermore, in the DiT architecture diffusion model, motion dynamics from reference videos can be injected in a unified token format, rather than through separate spatial, prompt-based, and temporal injections.
However, in reference videos, foreground and background, as well as the appearance and motion of subjects, are often coupled together. This results in redundant information in the reference data, which can negatively impact video generation. Therefore, decoupling the foreground from the background, as well as separating the appearance information of moving objects from their motion, remains challenge.

In this paper, we introduce a novel zero-shot Leveraging Motion Prior (LMP) framework, for DiT video diffusion models, enabling users to control video generation using motion from reference videos. To this end, we first analyze the attention mechanism in the DiT architecture and design a foreground-background disentangle module. This module disentangles foreground objects from the background in reference videos, ensuring the background does not affect the generated video. Next, we develop a motion transfer module to inject tokens of separated foreground objects from the reference video into the generation process, allowing the subject in the generated video to move according to the motion of the reference object. However, tokens from the reference video often carry appearance information, which can interfere with the generation of the subject video. To address this, we design an appearance separation module to remove subject information from the reference video tokens. Through these modules, our framework enables motion transfer in DiT models without any training. To evaluate our method, we annotate a video dataset with detailed prompts and design evaluation metrics for motion transfer and subject appearance preservation. Extensive experiments demonstrate the effectiveness of our approach.


Our main innovations are as follows:
\begin{itemize}
\item We are the first to propose a zero-shot framework LMP for DiT motion transfer in both text-to-video and image-to-video generation. Built upon a pre-trained DiT video diffusion model, our framework enables the transfer of motion from a reference video to the generated video without any training, making it plug-and-play.
\item Our fore-background disentangle module leverages attention mechanisms to separate foreground objects from the background. Our reweighted motion transfer module enables the generated video to reference the extracted foreground information. Finally, our appearance separation module removes the subject's appearance information from the reference video, retaining only the motion details.
\item We conduct detailed prompt annotations on the DAVIS video dataset and design evaluation metrics for motion transfer and subject appearance preservation. Extensive qualitative and quantitative experiments demonstrate the effectiveness of our method.
\end{itemize}

\section{Related Works}
\subsection{Video Diffusion Models}
As image diffusion models have flourished and demonstrated impressive results, the body of work related to video diffusion has gradually expanded. 
Researchers and developers are leveraging the principles of image diffusion models to extend their capabilities into the dynamic and temporally complex domain of video generation. 
In the realm of text-to-video (T2V) synthesis, the field has witnessed several pioneering approaches. 
Initial endeavors such as Imagen Video \cite{ho2022imagen} and Make-A-Video \cite{singer2022make} tackled T2V generation at the pixel level, leading to notable advancements but also encountering constraints in video duration and quality due to computational intensity. 
To address these limitations, MagicVideo \cite{zhou2022magicvideo} emerged, introducing a novel autoencoder trained specifically on video data, which, akin to the impact of Latent Diffusion Models (LDMs) \cite{rombach2022high} in the text-to-image (T2I) sector, significantly improved the computational efficiency of T2V tasks.
In the current landscape, there exists an abundance of generalizable open-domain text-to-video generation models capable of producing high-definition videos with dynamic visual fidelity \cite{Mullan_Hotshot-XL_2023, Sterling2023ZeroScope,chen2023videocrafter1,wang2023modelscope}. 
Recently, with the emergence of diffusion transformer (DiT) architectures \cite{peebles2023scalable}, a number of video generation models based on DiT \cite{kong2024hunyuanvideo, wang2025wan, yang2024cogvideox} have appeared. Compared to models based on the Unet architecture, DiT video generation models can produce videos of higher clarity, longer duration, and richer semantics. However, these models often control motion solely through prompts, making it difficult to achieve finer control over motion.

\begin{figure}[t]
  \centering
  \includegraphics[width=\linewidth]{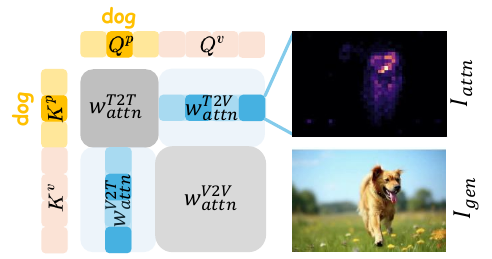}
  \caption{The core idea of our FBDM. We achieve disentanglement of the foreground and background by utilizing text-video attention maps and video-text attention maps.}
  \label{fig:FBDM}
\end{figure}

\subsection{Controllable Motion Generation}
To address the limitations of using only prompts to control motion in video diffusion models, a large body of work \cite{zhang2023adding, zhao2024uni,ma2024follow} aims to ensure that the generation results align with given explicit control signals, such as depth maps, human poses, optical flows, etc. These methods have significant limitations in control, as they can only manage specific positions or poses, rather than the dynamic patterns of motion. Additionally, the control medium is difficult for users to obtain.
Another set of training-based models \cite{he2022latent, wang2023videocomposer, chen2023control, zhao2024motiondirector, wang2024motionctrl,Esser2023StructureAC, dai2023animateanything, Yin2023DragNUWAFC} has been developed, incorporating an additional branch to accept control signals. These approaches require substantial training costs and are constrained by data bias during training, which can lead to decreased performance on unseen data. 
To address these issues, a series of works \cite{Xiao2024VideoDM, Yatim2023SpaceTimeDF, Geyer2023TokenFlowCD, chen2024motion, meral2024motionflow, wang2024motioninversionvideocustomization} have explored the latent properties of video diffusion models, achieving zero-shot motion control. However, current works are based on the Unet architecture, with little exploration of the latest DiT architecture.

\begin{figure*}[t]
  \centering
  \includegraphics[width=\linewidth]{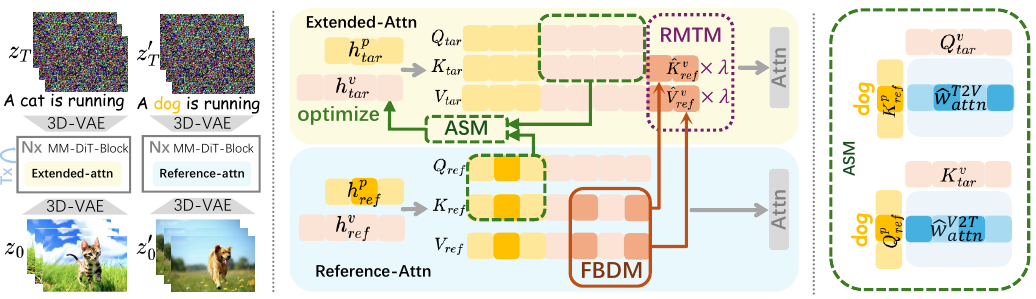}
  \caption{The pipeline of our LMP framework in each MM-DiT block. For the first $T_1$ denoising steps, we use FBDM and RMTM to enable the target video to reference the motion from the reference video. For denoising steps $T_2$ to $T_3$, we employ ASM to suppress the subject appearance information of the reference video in the target video.}
  \label{fig:pipeline}
\end{figure*}

\subsection{Attention Operations in Diffusion Model}
In diffusion models, attention operations often carry explicit semantics, and many existing works utilize these operations for various tasks. In Unet, self-attention can be used for style transfer and appearance reference \cite{yang2023reco, Deng2024ZZS},
cross-attention enables position-specific generation \cite{chen2024motion, jain2024peekaboo}, and temporal-attention ensures motion consistency.
Some recent works \cite{chen2023control, Ma2023TrailBlazerTC}. 
\cite{Zhou2025FreeBlendAC,Yu2025OmniPaintMO,Zhu2025KVEditTI,Tewel2024AdditTO} have explored manipulating the self-attention mechanisms in DiT for image-editing tasks. 
DitCtrl \cite{Cai2024DiTCtrlEA} leverages the attention mechanism in DiT to maintain a consistent subject when generating long videos.
There is a work \cite{Pondaven2024VideoMT} similar to ours that uses AMF to capture the motion trends of objects in reference videos, yet it overlooks the reference of motion poses and is not available for image-to-video control. Our work, by contrast, capitalizes on the global view of tokens in DiT, enabling the target video to fully reference the motion trends, poses, and styles of the reference video. Our method is capable of motion control generation for both text-to-video and image-to-video settings.

\section{PRELIMINARIES}
In this section, we briefly revisit the fundamental concepts of video diffusion models and the task of motion transfer.

\subsection{Video Diffusion Model}
Video Diffusion models are designed to produce high-quality, diverse videos. To reduce computational costs, Rombach et al. \cite{rombach2022high} proposed a Latent Diffusion Model (LDM) that conducts the denoising process in a latent space. This model features a Variational Autoencoder (VAE) with an encoder $\mathcal{E}$ to condense the original pixel to latent space, and a decoder $\mathcal{D}$ to revert from latent to pixel space. 
The training of the video diffusion model hinges on a noise prediction loss function:
\begin{equation}
    \mathcal{L} = \mathbb{E}_{\mathbf{z}_0,c,\epsilon\sim \mathcal{N}(0,I),t}[||\epsilon-\epsilon_{\varphi}(\mathbf{z}_t,t,\mathbf{c})||^2_2],
\end{equation}
where $z_0$ is the latent code of the training sample, \textbf{c} is the condition, $\epsilon$ is the Gaussian noise, and $t$ is the time step. The noised latent code $\mathbf{z}_t$ is determined as:
\begin{equation}
    \mathbf{z}_t = \sqrt{\overline{a}_t}\mathbf{z}_0+\sqrt{1-\overline{a}_t}\mathbf{\epsilon},\overline{a}_t = \prod^t_{i=1}a_t,
    \label{eq_z_denoise}
\end{equation}
where $a_t$ is a hyper-parameter used for controlling the noise strength based on time $t$.

The DiT architecture \cite{yang2024cogvideox} is a novel diffusion architecture that employs a transformer as its denoising network. Unlike the previous UNet architecture, it does not differentiate between self-attention, cross-attention, and temporal attention. Instead, it processes concatenated sequences of textual-prompt and image-patch tokens through unified MM-DiT blocks. Each block computes attention based on the concatenated tokens as follows:
\begin{equation}
\begin{aligned}
    A &= softmax([Q^p,Q^v][K^p, K^v]^\top/\sqrt{d_k}), \\
    h &= A \cdot [V^p, V^v],
    \label{eq:base}
\end{aligned}
\end{equation}
where \( Q^p \) and \( Q^v \) are the queries for the target prompt and video patch, \(K^p \) and \( K^v \) are the keys for the prompt and video patch, \( V^p \) and \(V^v \) are the values for the prompt and video patch. \( A \) is the attention map, and \( h \) is the hidden states.

\subsection{Video Generation with Motion Transfer}
Based on video diffusion, the task of motion transfer is to refer to the motion of a given reference video $\mathcal{V}_{ref}$ and make the motion of the generated video $\mathcal{V}_{tar}$ similar to that in the reference video $\mathcal{V}_{ref}$.
The optimization objective can be formulated as follows:
\begin{equation}
    \mathcal{L} = \mathcal{E}_{\mathbf{z}_0,c,\epsilon\sim N(0,I),t}[||\epsilon-\epsilon_{\theta}(\mathbf{z}_t,t,\mathbf{c},\mathcal{M}_{ref})||^2_2],
\end{equation}
where \( \mathcal{M}_{ref} \) is the motion information extracted from \( \mathcal{V}_{ref} \). In the case of text-to-video generation, \( \mathbf{c} \) refers to the prompt. In the case of image-to-video generation, \( \mathbf{c} \) refers to both the reference image and the prompt.

\section{Methods}
\subsection{Overview}
In this section, we provide a detailed explanation of our proposed LMP framework. First, we analyze the attention maps of MM-DiT and utilize their properties to disentangle the foreground and background in the reference video. Then, we ensure the target video can reference the subject's motion from the reference video by leveraging the properties of the target video's attention maps, and we perform a reweighting operation on the reference video's tokens to balance their weights relative to the target video. To prevent the subject's appearance in the reference video from affecting the generated target video, we suppress the influence of the reference video's subject appearance in the target video. Finally, to enhance the practicality of our method, we describe how to apply real reference videos to our LMP. The pipeline is shown in Fig.\ref{fig:pipeline}.

\subsection{Fore-Background Disentangle Module}
In previous work based on the UNet architecture, a straightforward idea for the separation of foreground and background is to utilize cross-attention layers to identify regions in each frame with higher values in attention map corresponding to the foreground prompt. However, due to the absence of cross attention layers in the DiT architecture, which encodes spatio-temporal information into unified tokens instead, these prior methods cannot be directly applied any more. Nevertheless, we can draw inspiration from the idea of using attention map values.

First, we analyze the attention maps within the MM-DiT blocks of the DiT architecture. As illustrated in the fig. \ref{fig:FBDM}, the attention map is divided into four parts: text-text attention, text-video attention, video-text attention, and video-video attention, then we visualize the text-video and video-text attention maps. Specifically, we transpose and sum the values of the text-video and video-text attention maps, then average the values across all layers and attention heads. Subsequently, specific rows and columns corresponding to the subject words in the prompt are selected and reshaped into the form of \( f \times h \times w \), where $f$ is the latent frame number, $h$ and $w$ are the height and width of the latent. It could be observed that, the text-video and video-text attention maps effectively capture the relationship between the prompt and the generated video. This forms the basis of our foreground-background separation module.

To elaborate, we extract the subject words from the prompt of the reference video using an LLM, and then identify the tokens corresponding to high responded attention values in the text-video and video-text attention maps for the prompt tokens, defining these tokens as foreground tokens. Using this module, we can dynamically obtain the foreground tokens during the denoising process in real-time, without any additional computational or time overhead.



\subsection{Reweighted Motion Transfer Module}
After the Fore-Background Disentangle Module, we could obtain the foreground tokens from the reference video \( \mathcal{V}_{ref} \). Unlike the UNet architecture, which separates spatiotemporal information, DiT's each token  independently possesses spatiotemporal information and share a global view during attention operations. To enable the generated video \( \mathcal{V}_{tar} \) to reference the motion from \( \mathcal{V}_{ref} \), only the tokens of \( \mathcal{V}_{tar} \) are allowed to access the keys and values of reference tokens during the attention process.

More formally, we define three sources of Key-Value (KV) information: the prompt \( p \), the target video \( \mathcal{V}_{tar} \), and the reference motion tokens. To synchronously obtain the reference motion tokens, we denoise \( \mathcal{V}_{ref} \) in parallel with the target video $\mathcal{V}_{tar}$, then concatenate the keys and values of the reference motion tokens into the self-attention blocks, extending the formula \ref{eq:base} as follows:
\begin{equation}
\begin{aligned}
    A &= softmax([Q^p_{tar},Q^v_{tar}][K^p_{tar}, K^v_{tar}, \hat{K}^v_{ref}]^\top/\sqrt{d_k}), \\
    h &= A \cdot [V^p_{tar}, V^v_{tar}, \hat{V}^v_{ref}],
\end{aligned}
\label{eq:trans}
\end{equation}
where $Q^p_{tar}$ and $K^p_{tar}$ are the queries and keys generated from target prompt. \(\hat{K}^v_{ref} \) and \( \hat{V}^v_{ref} \) are the keys and values generated from the reference motion tokens.

However, simply concatenating the keys and values of the reference video can cause the appearance of the reference video to influence the target video generation, overshadowing the effects of the prompt and initial frame. Therefore, we design a Reweighted Motion Transfer Module, which reweight the keys of the reference video. We updated the formula as follows:
\begin{equation}
\begin{aligned}
    A &= softmax([Q^p_{tar},Q^v_{tar}][K^p_{tar}, \lambda \cdot  \hat{K}^v_{ref}]^\top/\sqrt{d_k}), \\
    h &= A \cdot [V^p_{tar}, V^v_{tar}, \hat{V}^v_{ref}],
\end{aligned}
\label{eq:rew}
\end{equation}
where \( \lambda \) is the reweighting coefficient. After the reweighting operation, we can approximately balance the target video's reference to the motion of the reference video while minimizing interference from the appearance of the subject in the target video.

\begin{algorithm}[tb]
    \caption{LMP Algorithm}
    \label{alg:init}
    \begin{algorithmic}[1] 
    \STATE \textbf{Input}: Reference video $\mathcal{V}_{ref}$, target video prompt $\mathbf{c}$
    \STATE Reference video prompt $\mathbf{c}' \leftarrow \mathbf{LLM}(\mathcal{V}_{ref})$
    \STATE Initialize $z_T \sim \mathcal{N}(0,\mathcal{I})$
    \STATE $h_{tar} \leftarrow \mathbf{Mapping}(z_T, \mathbf{c})$
    \FOR{all t=$T$, $T-1$, ... 0} 
        \STATE $z'_t \leftarrow RVMT(\mathcal{V}_{ref},t)$ \textcolor{gray}{$\triangleright$ Obtain $z'_t$ from eq.\ref{eq:real}}
        \STATE $h_{ref} \leftarrow \mathbf{Mapping}(z_t', \mathbf{c}')$
        \FOR{all n=1,2,...,N} 
        \STATE \textcolor{gray}{$\triangleright$ N MM-DiT blocks}
            \IF{$T_3 < t < T_2$}
                \STATE $h_{tar} \leftarrow \mathbf{ASM}(h_{tar}, h_{ref}, \mathbf{c})$ \textcolor{gray}{$\triangleright$ Update $h_{tar}$ by eq.\ref{eq:7},\ref{eq:8},\ref{eq:9}}
            \ENDIF
            \IF{t > $T_1$}
                \STATE $h_{tar} \leftarrow \mathbf{RMTM}(\mathbf{FBDM}(h_{ref}, \mathbf{c}'), h_{tar}) $
                \STATE $h_{tar} \leftarrow \mathbf{REST}(h_{tar}) $ \textcolor{gray}{$\triangleright$ Update $h_{tar}$ by the rest part of MM-DiT block}
            \ELSE
                \STATE $h_{tar} \leftarrow \mathbf{MMDiT}(h_{tar})$
            \ENDIF
            \STATE $h_{ref} \leftarrow \mathbf{MMDiT}(h_{ref})$
        \ENDFOR
    \ENDFOR
    \STATE $z_0 \leftarrow \mathbf{Projection}(h_{tar})$
    \STATE \textbf{output}: Denoised target latent $z_0$
\end{algorithmic}
\label{algo}
\end{algorithm}

\subsection{Appearance Separation Module}

With the Reweighted Motion Transfer Module, the motion from \( \mathcal{V}_{ref} \) can be involved into the target video. However, solely relying on the reweighting parameter to decouple motion and appearance information from the reference tokens is insufficient, as it only partially reduces the influence of appearance. Therefore, we introduce the Appearance Separation Module to explicitly remove the appearance influence of the subject from the reference video.

As established in Section 4.2, the interaction between prompts and generated videos is reflected in the attention maps. A straightforward approach is to reduce the text-video and video-text attention values corresponding to the subject prompts of the target and reference videos. Specifically, we first obtain the prompt tokens for each denoising step of the reference video and concatenate them with the tokens of the target video. We then compute the attention map as follows:
\begin{equation}
    A' = softmax([Q^p_{ref},Q^v_{tar}][K^p_{ref}, K^v_{tar}]^\top/\sqrt{d_k}),
    \label{eq:7}
\end{equation}
where $Q^p_{ref}$ and $K^p_{ref}$ are queries and keys generated from reference video prompts. Subsequently, we obtain the attention values of specific rows and columns in the text-video and video-text attention maps corresponding to the subject tokens in the reference prompt. The top 1/5 of these values are chosen for computing their average. The formula is as follows:
\begin{equation}
    \mathcal{L} = \frac{1}{k} \sum_{j=1}^{k} a_{(j)}, \quad where\  k = \left\lceil \frac{n}{5} \right\rceil,
    \label{eq:8}
\end{equation}
where $a$ is the selected attention value, n
is the total number of the attention values.
To achieve lower attention values for the subject prompt in the reference video, we aim to minimize the average attention loss \( L \). This is accomplished through gradient descent. Specifically, we freeze the parameters of the entire DiT model and apply gradient descent to the hidden states of the target video to optimize its parameters, as shown in the following formula:
\begin{equation}
    h' \leftarrow h - \beta \cdot \nabla {\mathcal{L}},
    \label{eq:9}
\end{equation}
where $h'$ is the optimized hidden states of the target video. $\beta$ is the learning rate. With this explicit suppression module for separating the reference subject, the appearance of the reference video's subject no longer affects the generation of the target video. Notably, the ASM is optional and is only necessary when the subjects in the reference and target videos are different (e.g., a dog vs. a cat). If the subjects are the same, this module can be omitted.

\begin{figure*}[t]
  \centering
  \includegraphics[width=\linewidth]{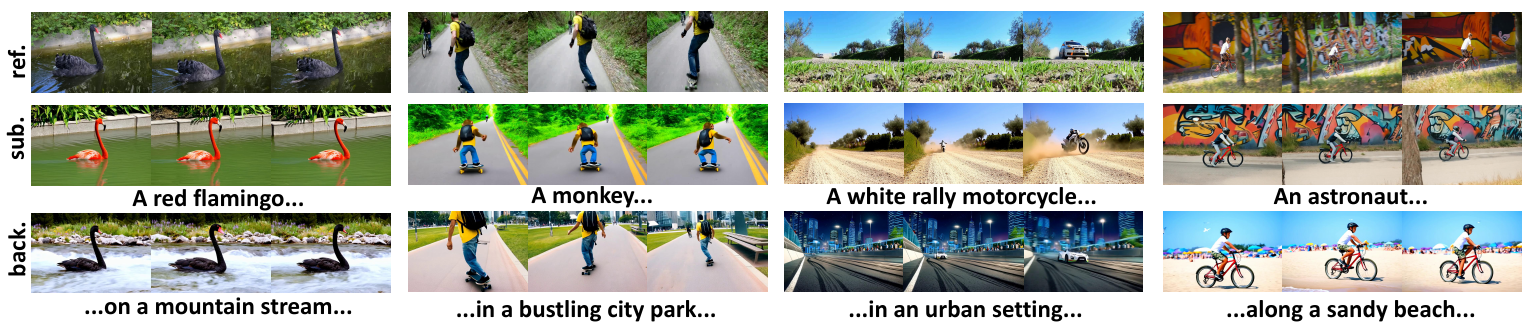}
  \caption{The text-to-video results of our LMP framework. The original videos are available in the supplementary material.}
  \label{fig:t2vresults}
\end{figure*}

\begin{figure*}[t]
  \centering
  \includegraphics[width=\linewidth]{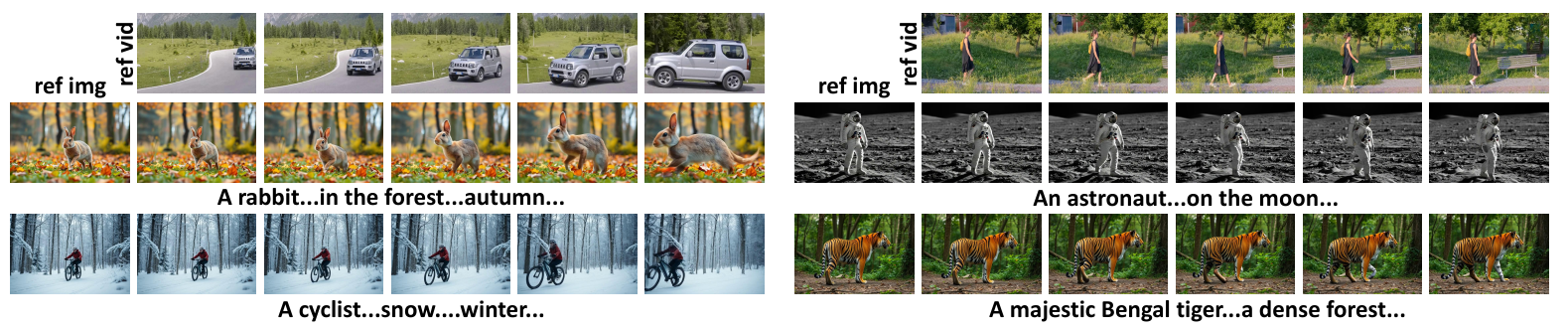}
  \caption{The image-to-video results of our LMP framework. The original videos are available in the supplementary material.}
  \label{fig:i2vresults}
\end{figure*}

\subsection{Real Video Motion Transfer}
In earlier sections, we explained how our method enables the target video to reference the reference video through simultaneous denoising. When the reference video is also generated, the process is straightforward. However, real-world applications often require real videos as references. Thus, we need a way to obtain initial noise for real videos to synchronize the denoising process.
While DDIM Inversion \cite{song2020denoising} is a popular method, it is designed for UNet architectures and performs poorly on diffusion transformers, with high time consumption. To address this, we propose a more efficient solution: directly adding noise to real reference videos proportionally, as shown in the following equation:
\begin{equation}
    z_t' = \lambda_t \cdot z_0' + (1-\lambda_t) \cdot \epsilon,
    \label{eq:real}
\end{equation}
\(\lambda_t\) represents the degree of noise added. One advantage of this approach is, when \( t = 0 \), \( z_t \) becomes \( z_0 \), making the entire denoising trajectory controllable. This ensures that the denoising process starts from a known initial noise state, allowing for precise control over the trajectory.
Our LMP method can reference both generated and real reference videos, enhancing its practical value. The LMP algorithm is shown in alg. \ref{alg:init}

\section{Experiments}
\subsection{Experimental Setups}
\textbf{Inference Setting.} 
We utilize CogVideoX-5B \cite{yang2024cogvideox} as our video generation base model, whose parameters remain frozen. Our algorithm is fully implemented during the inference stage, thus it does not require any training. 
We use qwen2.5-VL-72B-instruct \cite{qwen} as our LLM to generate dense prompts.
The total denoising timestep $T$ is set to 50.
The hyper-parameter $T_1$ is set to 40, $T_2$ is set to 45, and $T_3$ is set to 35.
The reweighting parameter is set to 0.98. The learning rate $\beta$ is set to 100.
The height and width of the generated video are 480 and 720. Each video has 49 frames.
All our experiments are conducted on a single NVIDIA H20 GPU.

\textbf{Dataset and metrics.}
Following \cite{Yatim2023SpaceTimeDF, Pondaven2024VideoMT},
we use the DAVIS dataset \cite{Pont-Tuset_arXiv_2017} to evaluate our method, which is originally for semantic segmentation and lack prompt annotations. We employ GPT to annotate it comprehensively from perspectives like subjects, motion, backgrounds, and overall style. The dataset and prompts are combined into three parts: one with original video and annotations for motion transfer testing, one with modified subject annotations to check reference subject suppression, and one with altered background prompts to verify foreground-background separation.
We use four metrics to evaluate our method: image-text alignment (Align), frame consistency (Cons.), PickScore (Pick.) \cite{kirstain2024pick}, and Motion Fidelity (MF). The image-text alignment metric assesses frame-by-frame consistency with the prompt using the CLIP score. The frame consistency metric uses the CLIP Score to measure the similarity between frames, reflecting the temporal consistency of video generation. The PickScore employs a pre-trained network to evaluate the quality of the generated frames. The MF metric evaluates motion consistency with the reference video by leveraging a pre-trained tracker to monitor subject movement.

\begin{figure*}[t]
  \centering
  \includegraphics[width=\linewidth]{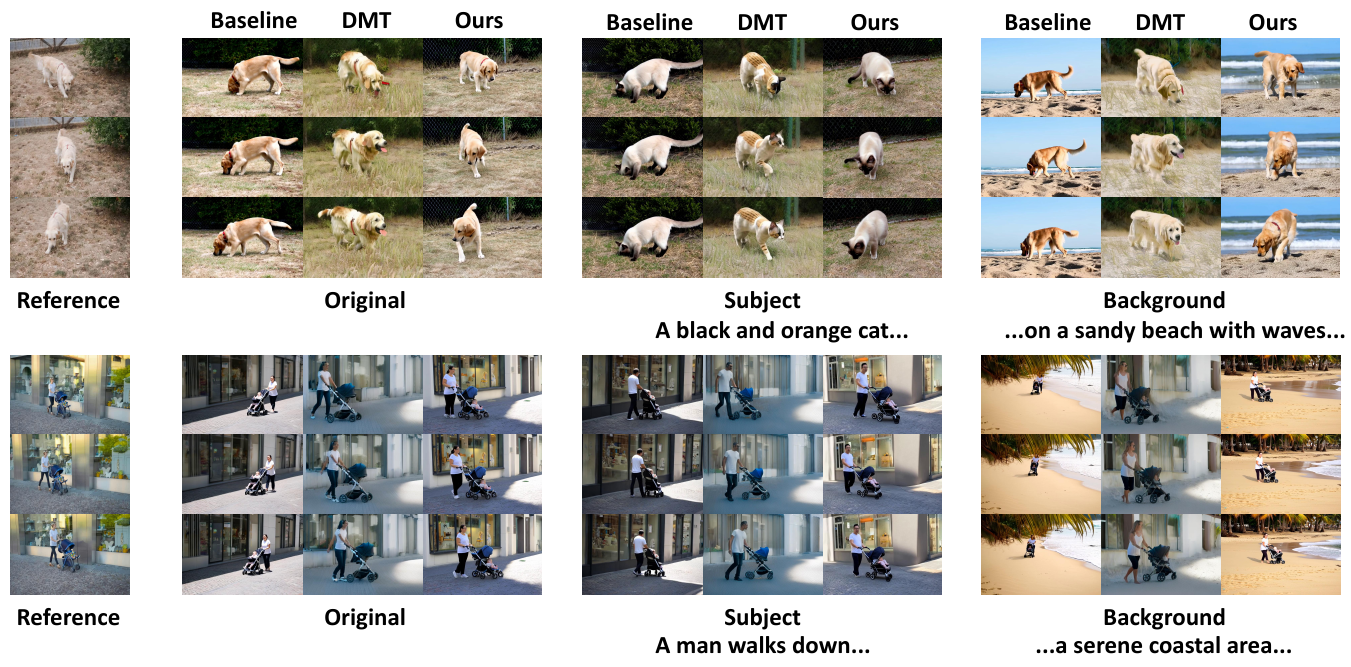}
  \caption{Quality comparison results on different methods in text-to-video setting. }
  \label{fig:compare}
\end{figure*}

\begin{figure}[t]
  \centering
  \includegraphics[width=\linewidth]{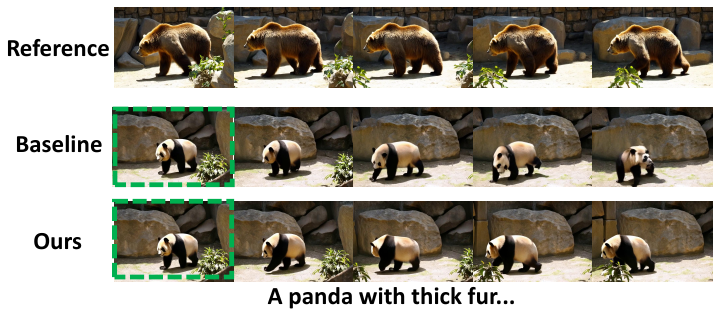}
  \caption{Quality comparison results on different methods in image-to-video setting. The green dashed box indicates the reference image.}
  \label{fig:img_cmp}
\end{figure}

\subsection{Results}
Fig. \ref{fig:t2vresults} visually demonstrates the effectiveness of our method in text-to-video setting. It contains four sets of data, where the first row of each set is the reference video, the second row shows the effect of changing the subject prompt while keeping the background prompt unchanged, and the third row illustrates the effect of modifying the background while keeping the subject unchanged. We can observe that our method effectively captures the motion patterns from the reference video and generates both the subject and modified background based on the prompts. For example, in the first set of data, we retain the motion of the black swan swimming to the right but can transform it into a flamingo or replace the background with a rapid mountain stream.

Different from previous motion-transfer or video-to-video methods that generate videos with structures identical to the reference video, our method imposes minimal manual supervision on the motion. Instead, we leverage DiT's strong generative capabilities to create videos with different overall structures but the same subject motion patterns as the reference video. As shown in Fig. \ref{fig:i2vresults}, the rabbit in our generated video starts at a different position and size than the car in the reference video but follows the car's motion pattern. Similarly, the astronaut is generated with the same motion as the reference video despite a different initial position. This demonstrates our successful decoupling of motion from video structure, proving our method's effectiveness and potential for image-to-video tasks.

\subsection{Comparison with Other Methods}
\label{sec:sota}
To verify the effectiveness of our method, we conducted both quantitative and qualitative experiments. We compared our proposed LMP method with the baseline method, CogVideoX \cite{yang2024cogvideox}. Since there are no other open-source motion transfer methods based on the DiT architecture currently available, we compared our method with the previous zero-shot SOTA method, DMT \cite{Yatim2023SpaceTimeDF}, which is based on the U-Net architecture. It is important to note that due to limitations imposed by the U-Net-based model, ZeroScope \cite{Sterling2023ZeroScope}, DMT can only generate videos with 24 frames at a resolution of 576×320. To ensure a fair comparison, we also extracted the first 24 frames of the videos generated by our method and reshaped them to the same resolution.

\noindent \textbf{Qualitative analysis.}
As shown in fig. \ref{fig:compare}, we present the results of the qualitative experiments in text-to-video setting, comparing our method with the baseline and the previous SOTA method (DMT) across two sets of data.  
The baseline method completely fails to capture the motion from the reference video. Notably, in the experiments, we used the same initial seed, so the generated videos exhibit similarities when the subject or background in the prompt is altered.  
While DMT initially mimics the motion from the reference video, it diverges in subsequent frames and shows no response to prompts involving background changes. This may be due to its inability to decouple motion from the background in the reference video.  
In contrast, our method effectively references the motion from the video and responds well to prompt edits, demonstrating superior performance in both motion transfer and prompt-based content adaptation.
As shown in fig. \ref{fig:img_cmp}, we compare the generation capabilities of our LMP framework under the image-to-video setting. Compared to the baseline method, where the panda moves randomly, our approach enables the panda to move according to the motion of the bear in the reference video.

\noindent \textbf{Quantitative analysis.}
Table 1 presents the quantitative comparison of our method with other methods. In terms of the image-text alignment metric, we observe that the performance of our method is comparable to the baseline method. This indicates that our method can generate the specified subject appearance and background according to the prompt while referencing the motion prior from the reference video. On the other hand, the DMT method shows a lack of prompt consistency in background changes, suggesting that DMT does not effectively decouple the background and the motion subject from the reference video.
For the frame consistency and PickScore metrics, our method outperforms DMT. This is because our method is adapted to the DiT architecture, whereas the DMT method is limited to the U-Net architecture. The DiT architecture's foundational model exhibits superior performance in video generation compared to the U-Net architecture.
Regarding Motion Fidelity, our method demonstrates better motion transfer capabilities than DMT, effectively referencing the motion from the reference video. These metrics collectively showcase that our method achieves state-of-the-art (SOTA) performance in video generation quality, prompt consistency, and the ability to reference motion.

\begin{table}[t]
\caption{Quantity results on baseline method and SOTA method and ours.  Higher values indicate better performance.}
    \centering
    \begin{tabular}{cccccc}
        \toprule
                &  &  Align & Cons. & Pick. & MF \\
        \midrule
        \multirow{3}{*}{Ori.} & Baseline   & \textbf{23.21}  & 0.93 & 20.52 & 0.33\\
                                  & DMT       & 22.31  & 0.95  & 18.55 & 0.85 \\
                                  & Ours       & 23.12  & \textbf{0.96} & \textbf{20.51} & \textbf{0.93}\\
                                  \hline
        \multirow{3}{*}{Sub.} & baseline   & 25.76   & \textbf{0.96}  & 20.69 & 0.33 \\
                                  & DMT       & 25.76   & 0.92 & 19.29  & 0.83 \\
                                  & Ours       & \textbf{27.47} & \textbf{0.96}  & \textbf{20.49} & \textbf{0.88} \\
                                  \hline
        \multirow{3}{*}{Back.} & Baseline   & 25.54   & 0.95 & 20.16 & 0.32\\
                                  & DMT       & 22.54   & 0.95 & 19.00 & 0.86 \\
                                  & Ours       & \textbf{26.46}   & \textbf{0.97} & \textbf{21.06} & \textbf{0.90} \\    
                                  \hline
                                  \hline
        \multirow{3}{*}{Mean} & Baseline   & 24.84   & 0.95 & 20.46 & 0.33 \\
                                  & DMT       & 24.30  & 0.94 & 18.95 & 0.85 \\
                                  & Ours       & \textbf{25.68}   & \textbf{0.96} & \textbf{20.69} & \textbf{0.90} \\                             
        \bottomrule
    \end{tabular}
    
    \label{tab:metric}
\end{table}


\subsection{Ablation Study}
To verify the effectiveness of each module in our method, we conducted both qualitative and quantitative analyses. Since the RMTM module is the cornerstone of our motion transfer method and removing it would revert to the baseline method, we only performed ablation studies on the FBDM and ASM modules.

\begin{figure}[t]
  \centering
  \includegraphics[width=0.9\linewidth]{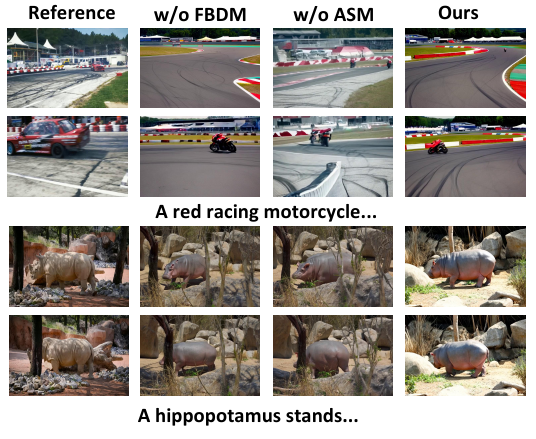}
  \caption{Quality results of ablation studies. The original videos are available in supplementary material.}
  \label{fig:abl}
\end{figure}

\noindent\textbf{Impact of FBDM.}
The FBDM leverages the properties of attention maps in the MM-DiT block to distinguish between the moving subject and background in the reference video, thereby preventing background interference in the target video. As shown in the upper part of Fig. \ref{fig:abl}, removing this module causes the motorcycle to disappear due to background interference. In the lower part, the hippopotamus gains an extra ear. Tab. \ref{tab:abl} indicates that without the FBDM, the generated video is disrupted by additional background information, reducing text consistency and introducing redundant details. Moreover, the reduced emphasis on motion decreases control accuracy.

\noindent\textbf{Impact of ASM.}
The ASM is designed to remove the appearance information of the subject from the reference video, preventing interference with the target video's subject concept. This module is only used when the subject in the target video differs from that in the reference video. As shown in Fig. \ref{fig:abl}, removing this module allows the appearance of the car to interfere with the motorcycle, resulting in either two motorcycles or a distorted, wide motorcycle. The rhino's body also becomes distorted. Tab. \ref{tab:abl} shows that without the ASM, the quality of the generated video declines due to interference from the reference subject's appearance. Additionally, consistency with the prompt decreases, and the ability to reference motion is weakened by the disturbance of appearance information.

    

\begin{figure}[tbp]
  \centering
  \includegraphics[width=0.9\linewidth]{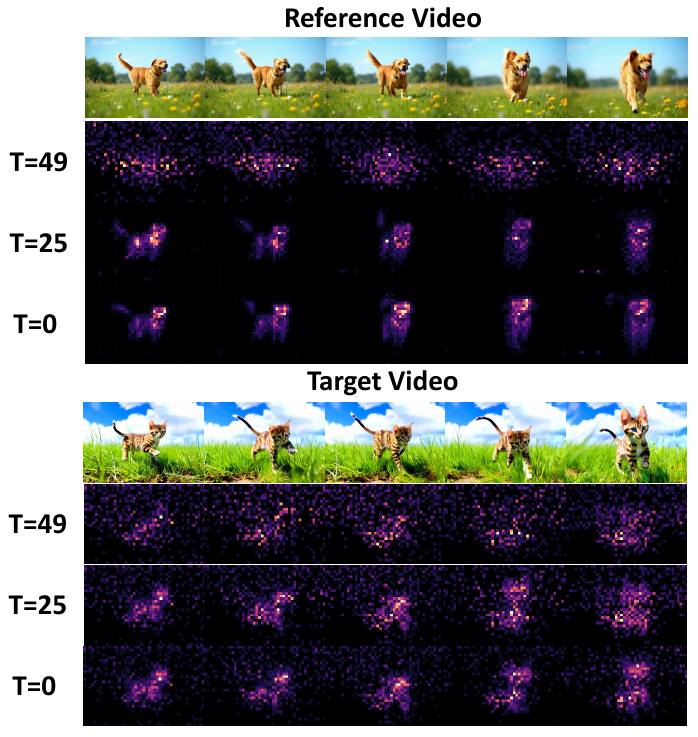}
      \captionof{figure}{Attention maps of reference video and target video. }
      \label{fig:attn}
\end{figure}

\begin{table}[t]
\caption{The quantity results of ablation study.}
    \centering
    \begin{tabular}{ccccc}
        \toprule
                & Align & Cons. & Pick. & MF \\
        \midrule
                w/o FBDM   & 23.98  & 0.93  & 19.46 & 0.77 \\
                w/o ASM  & 23.47  & 0.93 & 19.13 & 0.82  \\
                Ours  & \textbf{25.68}   & \textbf{0.96} & \textbf{20.69} & \textbf{0.90}  \\
        \bottomrule
    \end{tabular}
    
    \label{tab:abl}
\end{table}


\subsection{Visualization of Attention Maps}
In this section, we demonstrate the effectiveness of our Fore-Background Disentangle Module and Reweighted Motion Transfer Module by visualizing the attention maps of the reference and target videos. As shown in the fig. \ref{fig:attn}, the upper part displays the attention map of the subject dog in the real video after converting it to noise and denoising using eq. \ref{eq:real}. The resulting attention map is relatively clean, especially at T=25, allowing easy separation of foreground and background to extract the subject using the Fore-Background Disentangle Module.
In the lower part, after applying the Reweighted Motion Transfer Module, the attention map of the target video's subject (a cat) becomes similar to that of the reference video. As denoising progresses, the attention map gradually becomes cleaner. This similarity in attention maps confirms that the motion of the cat in the target video aligns with the motion of the dog in the reference video, validating the effectiveness of our LMP method.

\section{Conclusion}
In this paper, we introduced LMP, a zero-shot framework enabling DiT-based video generation models to reference motion from a given video. Unlike the three-layer attention mechanism in U-Net-based models, DiT features a single self-attention layer. We analyzed this layer and developed a Fore-Background Disentangle Module to decouple the foreground and background in the reference video. Additionally, we proposed a Reweighted Motion Transfer Module to allow motion reference from the input video to the target video, with a reweighting mechanism to balance its impact. To prevent the reference video's subject appearance from affecting the target video, we designed an Appearance Separation Module using gradient descent to eliminate such influence. Furthermore, we applied direct noise addition to adapt our method for real videos as references. We also utilized an LLM to annotate a video dataset with dense prompts and designed evaluation metrics. Extensive experiments demonstrate the effectiveness of our approach.
\bibliographystyle{ACM-Reference-Format}
\bibliography{acmart}

\end{document}